\documentclass[10pt,twocolumn,letterpaper]{article}

\usepackage[pagenumbers]{cvpr} 

\usepackage[table,xcdraw]{xcolor}

\usepackage{times}
\usepackage{epsfig}
\usepackage{graphicx}
\usepackage{amsmath}
\usepackage{amssymb}
\usepackage{pifont}
\usepackage{stmaryrd}
\usepackage{trimclip}
\newcommand{\cmark}{\ding{51}}%
\newcommand{\xmark}{\ding{55}}%

\usepackage{booktabs}
\usepackage{caption}
\usepackage{multirow}
\usepackage{multicol}

\usepackage[pagebackref=true,breaklinks=true,letterpaper=true,colorlinks,bookmarks=false]{hyperref}
\definecolor{myRed}{RGB}{255, 45, 85}
\definecolor{myBlue}{RGB}{0, 122, 255}
\newcommand{\ann}[1]{{\color{black}{#1}}}

\newcommand{\saura}[1]{{\color{black}{#1}}}

\definecolor{cvprblue}{rgb}{0.21,0.49,0.74}

\def\paperID{3} 
\def\confName{3DV\xspace}
\def\confYear{2024\xspace}

\title{PersonalTailor: Personalizing 2D Pattern Design from 3D Garment Point Clouds}

\author{Sauradip Nag$^{1}$
\and 
Anran Qi$^{2}$
\and
Xiatian Zhu$^{3,4}$
\and
Ariel Shamir$^{5}$
\and \newline
{\small $^1$ Independent Researcher} ~
{\small $^2$ University of Tokyo, Japan} ~ 
{\small $^3$ Surrey Institute for People-Centred Artificial Intelligence, UK} \\
{\small $^4$ CVSSP, University of Surrey, UK} ~
{\small $^5$ Reichman University, Israel}
}
\begin{document}
\twocolumn[{%
\renewcommand\twocolumn[1][]{#1}%
\maketitle
\vspace{-6mm}
\begin{center}
    \centering
    \captionsetup{type=figure}
    \includegraphics[width=0.95\linewidth]{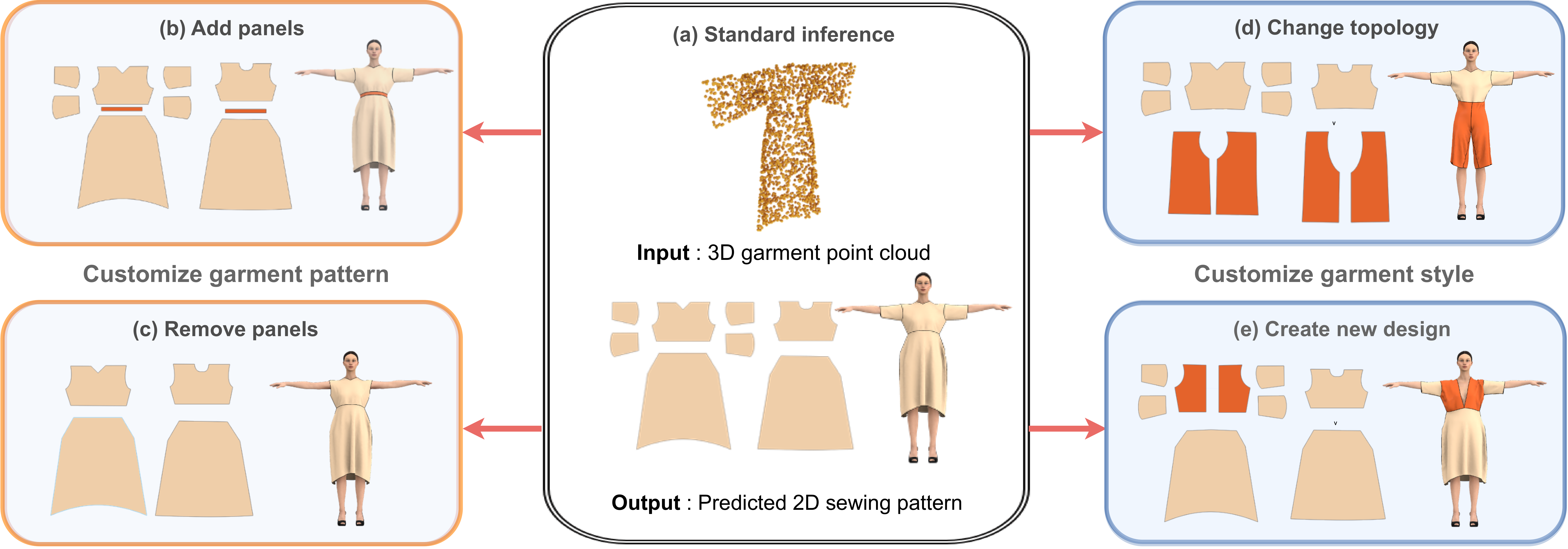}
    \captionof{figure}{\textbf{Illustration of personalized 2D pattern design.} Given a 3D point cloud, our goal is to predict the 2D panels and sewing pattern {\bf(a)}, and allow its \emph{personalization} based on a user text or sketch prompt. This supports various functionalities including {\bf(b)} adding new panels (\eg, waistbands), {\bf(c)} removing panels (\eg, sleeves), {\bf(d)} changing the topology (skirt \textrightarrow pant), and {\bf(e)} creating new design. }
   \label{fig:teaser}
\end{center}%
}]

\begin{abstract}
\vspace{-0.1in}
Garment pattern design aims to convert a 3D garment to the corresponding 2D panels and their sewing structure. 
Existing methods rely either on template fitting with  heuristics and prior assumptions,
or on model learning with complicated shape parameterization.
Importantly, both approaches do not allow for personalization of the output garment, which today has increasing demands.   
%
%
To fill this demand, we introduce {\bf\em PersonalTailor:} a personalized 2D pattern design method, where the user can input specific constraints or demands (in language or sketch) for personal 2D pattern design from 3D point clouds.
PersonalTailor first learns disentangled multi-modal panel embeddings based on unsupervised cross-modal association and attentive fusion. It then predicts 
2D binary panel masks individually using a transformer encoder-decoder framework.
%
Extensive experiments show that our PersonalTailor
excels on both personalized and standard garment pattern design tasks.
\end{abstract}    
\section{Introduction}

Converting a scanned model of a garment from a 3D point cloud to 2D panels and their sewing structure is termed \emph{pattern design}~\cite{pietroni2022computational,igarashi2005rigid,bartle2016physics}. Such a procedure allows simpler modeling, duplication and adjustments of existing garments in fashion,
because the 2D panels are necessary
for production with real fabric in the manufacturing process. 
Earlier works either modify predefined pattern templates to fit a particular body shape \cite{bartle2016physics,li2018foldsketch,yang2018physics,wang2018rule}, or flatten the surface subject to minimising surface cutting and flattening energy \cite{igarashi2005rigid,sharp2018variational,wang2003feature}.
To avoid such heuristics and rigid assumptions, recently learning based methods have emerged~\cite{korosteleva2022neuraltailor}, leveraging more annotated data samples~\cite{korosteleva2021generating}.
%
The key idea behind existing learning-based methods is to represent the garment shape as the garment sewing pattern (a set of panels) and predict each panel by regressing their hand-crafted edge features (\eg, edge vector and curvature coordinates). While being more general, such methods still lack the ability to personalize the resulting garment. 
%
However, personalization has increasing demand in modern retails \cite{FP} primarily as ready-made clothes cannot  satisfy the growing needs of individual customers.
%

In this work, we introduce a method for 
\emph{personalized} 2D pattern design with sketch and language.
The objective is to
infer the underlying 2D panels and their sewing structure from a 3D point cloud of a garment, following specific user-defined panel instructions (see Fig. \ref{fig:teaser}).
This is possible since a single garment could be converted by varying sewing structures with different topologies, varying panel numbers and shapes.
We choose language and sketches as the means of interaction
due to their simplicity and flexibility, whilst other forms
can be similarly integrated in our framework.
This problem is non-trivial
due to the extra need for fine-grained cross-modal alignment between language/sketch and point clouds,
along with the original 3D-to-2D mapping challenges.

To overcome these challenges, we propose 
{\bf\em PersonalTailor}, a novel method with two key components:
(1) A multi-modal panel embedding module 
that allows panel level association and information fusion across text/sketch and
point cloud;
(2) A binary panel mask prediction module
that decodes accurately the stitching panel masks.
Our personalization can be achieved 
by specifying flexible configurations of individual panel embeddings according to the user's prompt. 
\saura{In particular, our method supports a variety of garment styles, from simple frocks, tank tops, skirts, pants, and stylistic panel arrangements like skirt with belt to more elaborate complex jumpsuits (see Fig 1 and 4).
It supports also adding/removing panels, changing the topology, and creating a completely new design by personalizing the panel combinations.
With flexible visual prompt design,
it can even edit the length of dress or sleeves (Fig 5).}
%
The first module is based on unsupervised association between  point cloud local features and semantic features of user's prompt, and attentive fusion of the associated features,
yielding a semantic multi-modal panel embedding space.
Conditioned on the point cloud global representation, 
the second module then decodes each semantic embedding into
a binary panel 2D mask respectively in a transformer decoder design.
We further design  a StitchGraph module for 
more accurate stitch matching without exhaustive pairing.


Our {\bf contributions} are summarized as follows:
(I) We introduce the novel problem of personalizing 2D pattern design from 3D point clouds. (II) We propose {\bf\em PersonalTailor}, a method with two key components: {\em multi-modal panel embedding} for enabling panel-level personalization, and {\em binary panel mask prediction} for simple yet superior pattern design. PersonalTailor supports both standard and personalized garment pattern design. Extensive experiments show the superiority of our method over the state-of-the-art alternatives in both standard and personalization settings.


\section{Related works}

\noindent \textbf{2D pattern design} is a research problem with both geometry processing and physics-based simulation.
It has been studied in both graphics and vision communities.
Existing methods can be classified into three categories: 
(i) Modifying the predefined pattern templates to fit a particular body shape \cite{bartle2016physics,li2018foldsketch,yang2018physics,wang2018rule,wang2018learning}. 
(ii) Flattening the surface via minimizing the surface cutting and flattening energy \cite{igarashi2005rigid,sharp2018variational,wang2003feature}, fabrication-specific distortion energy \cite{pietroni2022computational}. 
(iii) Inferring the sewing pattern structure directly from 3D shape via deep learning methods \cite{korosteleva2022neuraltailor}. 
Our work belongs to the third category.
Compared to \cite{korosteleva2022neuraltailor},
the key difference is that 
our PersonalTailor uniquely allows panel personalization,
whilst also supporting the standard garment pattern design.
This is realized by
designing a multi-modal panel embedding module
and a panel mask prediction module.

\noindent\textbf{Transformers} was originally introduced as a sequence-to-sequence attention-based language model \cite{vaswani2017attention}.
It has been adopted over various computer vision tasks, \eg, image classification \cite{khan2022transformers, dosovitskiy2020image}, and 3D geometry learning \cite{zhao2021point}. 
There has been a popularity of vision-language models like CLIP \cite{radford2021learning} applying contrastive learning \cite{chen2020simple} across the features of a vision Transformer and a language Transformer. 
PointCLIP \cite{zhang2022pointclip} uses 2D image models to classify 3D point-clouds, but its performance is inferior when used as a pretrained backbone due to the domain gap between downstream task and pretraining data. 
Popular language models (\eg, CLIP) have been shown to be generally effective such as referring image segmentation \cite{jeyanathan2020immunological}.
Under this observation, here we extend the scope of language models for personalized garment pattern design.
To the best of our knowledge, this is the first work that predicts 2D masks from 3D point-cloud conditioned on language or sketch as prompt. 

\begin{figure*}[h]
    \centering
    \includegraphics[scale=0.21]{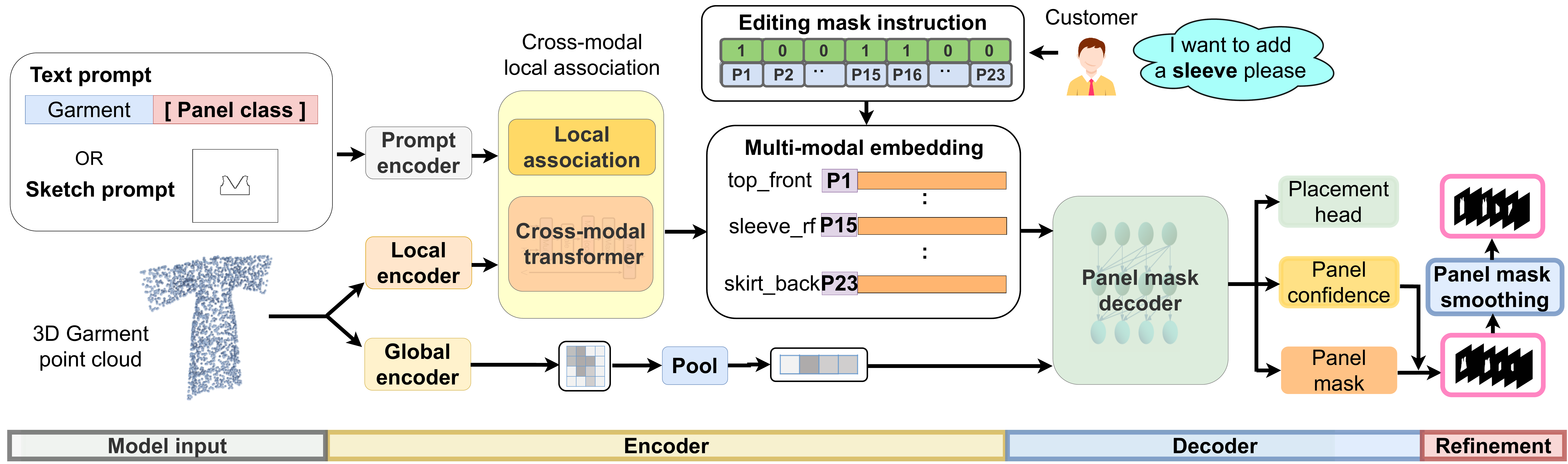}
    \caption{\textbf{Architecture of our PersonalTailor method}. Taking as input a 3D point cloud garment and the user's instruction (text prompt or sketch), (a) we extract local and global features for the point cloud, and the semantic feature for user input.
    (b) We then perform unsupervised  association 
    between local point cloud features and semantic features
    for cross-modal alignment and multi-modal fusion, leading to a disentangled multi-modal panel embedding space.
    (c)
    We further decode each multi-modal embedding to 
    a binary panel 2D mask, a confidence score and 3D placement.
    Positional encoding is applied for leveraging the spatial information of panels.
    (d)
    We refine and stitch the panel masks to obtain the final draped garment.
    }
    \label{fig:overview}
\end{figure*}

\noindent \textbf{Cross-modal association} 
methods could be grouped into two categories: single-stream and dual-stream. Single-stream models (\eg, VisualBERT \cite{li2019visualbert} and UNITER \cite{chen2020uniter})
concatenate the features of multiple modalities and then exploit a single  encoder for representation learning.
Instead, dual-stream models (\eg, CLIP \cite{radford2021learning} and CLIPasso \cite{vinker2022clipasso}) use a separate encoder for each modality before aligning them, with a flexibility of using modality-specific models and tackling more diverse downstream tasks. 
Typically, both approaches consider holistic alignment and information fusion.
However, our problem needs more fine-grained alignment at part level without the corresponding box annotation (\ie, weakly supervised). 
Object detection could be one intuitive solution \cite{li2020oscar}.
Nonetheless, there is a lack of strong pretrained garment panel detectors which 
demand a large of labeled training data.
To overcome this challenge,
we resort to unsupervised cross-modal association and fusion
in our model design.

\section{Method}
\label{sec:method}
\setlength{\belowdisplayskip}{5pt} \setlength{\belowdisplayshortskip}{2pt}

\setlength{\abovedisplayskip}{5pt} \setlength{\abovedisplayshortskip}{5pt}

\noindent \textbf{Overview:} 
Given a source 3D garment $G$ and the personalized requests $L$ from a customer, we aim to predict the corresponding 2D patterns (\saura{which can be stitched together to form} this garment) from a specific library of panels, 
their position in 3D, and their stitching structure.
Our key idea is to define an encoder-decoder network that could encode the multi-modal information from the point clouds and the user's instructions and decode it into the 2D panels structure.
An overview is depicted in Fig.~\ref{fig:overview}. 
\saura{We need to address two key challenges. First, as panel-level point-cloud segmentation is expensive to label, we want to learn the multimodal correspondence across the text/sketch and point cloud. Second, the multi-modal latent panel representations need to be disentangled from each other. Such a design allows us to directly manipulate the panel composition during editing, while visually mirroring these transformation results in producing the garments of different designs and topology as shown in Fig~\ref{fig:teaser}.}


To apply such learning based approach, we assume a given dataset $D=\{D_{train}, D_{test}\}$ including training and testing garment classes. 
Each subset is in the form of 
$\{P_{i}, L_{i}, Y_{i}\}_{i=1}^{N}$ where
$P_i$ is the garment point cloud,
$L_{i} = \{g_{i},p_{k=1}^{N_{i}}\}$ represents the garment class $g_{i}$ along with the related panel class labels $p_{k}$. 
$Y_{i} = \{m_{j=1}^{N_{i}}, v_{j=1}^{N_{i}}, c_{j=1}^{N_{i}}, r_{j=1}^{N_{i}}, t_{j=1}^{N_{i}}, s_{j=1}^{N_{i}}\}$ where $m_{j}$ is the 2D panel mask of this garment class, 
$v_{j}$  is an ordered list of 2D coordinates for the panel $m_{j}$, $c_{j}$ is the curvature of the edges for $m_{j}$, 
$r_{j}$ and $t_{j}$ are the rotation and  translation (Euler angles) used to define the panel 3D location around a particular body model, $s_{j}$ is the stitch information of $m_{j}$,
and $N_{i}$ is the total number of panels in a given garment $i$ design.  
More details on the ground-truth formulation are provided in supplementary file.
Note that, all the garment classes share the same set of panels,
and personalizing the panel configuration allows to form different garments.

\saura{Given the garment point-cloud  and the text/sketch panel prompts, each panel of the training data can be converted into a discrete feature map. However, directly combining local part level point-cloud features with its corresponding panel embedding is not trivial, since 
they lie in different feature spaces without part correspondence.
To align the point-clouds and the prompt embeddings of the garment, we propose to reduce the distribution gap  between the two modalities using optimal transport based cross-modal association. To further infuse the aligned point-cloud representation into the prompt embedding space, we perform cross-attention between the point-cloud features and the prompt features to obtain multi-modal information infused panel features.  
}

\begin{figure*}
    \centering
    \includegraphics[scale=0.18]{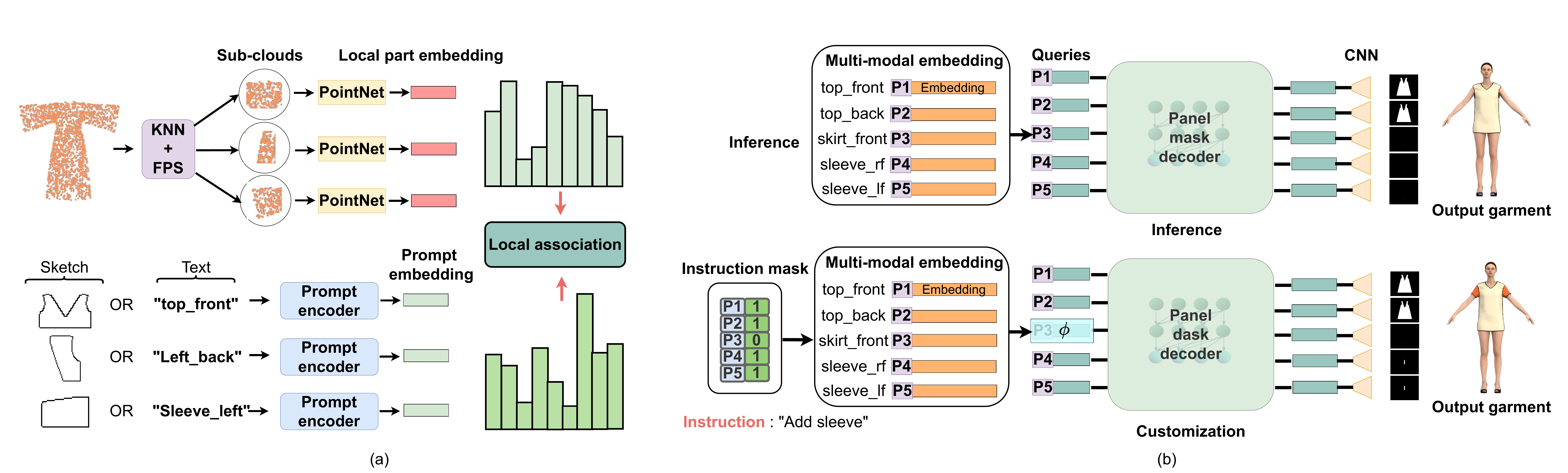}
    \caption{(a) {\bf\em Unsupervised cross-modal association} between point-cloud local representation and semantic representation of user's prompt. 
    (b) {\bf\em Binary panel mask decoder}: Except the standard inference, personalization can be flexibly supported, \eg, using an instruction mask.}
    \label{fig:decoder}
\end{figure*}

\subsection{Encoder: Multi-modal panel embedding}
\noindent {\bf Point-cloud local representation} \saura{Since a 3D garment is composed of multiple panels, it is useful to align the local patches of point-cloud (corresponding to panel or panel parts) with the panel prompts.} To facilitate panel extraction from 3D point-cloud (without panel-level annotation) in an unsupervised manner, we start by 
local patch analysis as in PointBERT \cite{yu2022point}.
%
Specifically, given a point cloud $P \in \mathbb{R}^{N \times 3}$, we first sample $g$ center points of $p$ via farthest point sampling (FPS). Then use $k$-nearest neighbor ($k$NN) to associate the nearest points in the point cloud to each center point, resulting in $g$ local patches $\{p_{i}\}_{i=1}^{g}$. We normalize the local patches by subtracting their center coordinates, disentangling the structure patterns and spatial coordinates. For efficiency, we employ a mini-PointNet to embed the patches $F_{loc} \in \mathbb{R}^{g \times D}$ with $D$ the embedding dimension.

\noindent {\bf Point-cloud global representation}
\saura{
We leverage the overall shape of garment as contextual information for estimating the sewing patterns.}
To that end
we use a PointTransformer ($\mathbb{T}$) \cite{zhao2021point} to extract the global representation of point cloud, as the self-attention has a global view field.
We first obtain per-point features $F_{p} = \mathbb{T}(P) \in \mathbb{R}^{N \times D}$ with $D$ the feature dimension, then apply positional encoding. 
We finally aggregate $F_{p}$ into a single feature vector $F_{global} = \phi(F_{p})$ by average pooling $\phi(.)$.

\paragraph{Personalization input representation} 
We consider (not limited to) both text prompt and sketch as the user instruction for personalization.
For text prompt, any prompt encoder (\eg, pre-trained CLIP\cite{radford2021learning} text encoder) can be used. \saura{We use a static prompt template of $Garment+ \{panel class\}$ where the panel class (denoted by $p_{i}$, $i \in \{1,..,K\}$) can be ``skirt-front'', ``jacket-sleeveless'' etc.}
For sketch, we use the SketchRNN \cite{ha2017neural}.
Formally, given a panel set $p$ required by the user, \saura{we obtain 512-D textual features (for textual prompt) for
$K$ panels from CLIP-encoder and the mean latent vector from Sketch-RNN's encoder 
for the
sketches of all the training garments belonging to $K$ panels. We then use a projection MLP to match the dimensions.}
We finally obtain the personalization feature $P_{loc} \in \mathbb{R}^{K \times D}$, where $K$ is the number of panels and $D$ is the feature dimension.

\subsubsection{Cross-modal local association}
Without ground-truth annotation of panels in a point cloud,
it is necessary to perform cross-modal association between local panel features
$F_{loc}$ and personalization feature $P_{loc}$ representation in an {\em unsupervised} manner.
The goal is to achieve multi-modal information fusion at the panel level
with both high-level semantic information from the personalization input (\eg, text prompt or sketch) and low-level fine-grained geometry information from the point cloud.

In essence, associating $F_{loc}$  and $P_{loc}$ is a set-to-set matching problem. 
Without pairwise labeling, we adopt the Wasserstein distance between the two sets of feature distributions (see Fig. \ref{fig:decoder}(a)). 
We define the cost of matching as the normalized mean-squared error (MSE) between $P_{loc}$ and $F_{loc}$ \cite{yu2022dual}. 
We denote the cost of moving $P_{loc}^{g}$ to $F_{loc}^{k}$ as $\delta_{g,k} = MSE(\hat{P}^{g}_{loc}, \hat{F}^{k}_{loc})$, where $\hat{P}$/$\hat{F}$ denotes an individual feature from $P$/$F$.
To encourage accurate local association, we minimize the following optimal transport cost:
\begin{align}
    OPT(P_{loc}^{g},F_{loc}^{k}) = \sum_{g=1}^{G}\sum_{k=1}^{K}f_{g,k}\delta_{g,k} \;\; \\ \text{where} 
    \sum_{g=1}^{G}\sum_{k=1}^{K}f_{g,k} = min(\sum_{g=1}^{G}w_{g}^{v},\sum_{k=1}^{K}w_{k}^{v})
\end{align}
where $w_{g}$/$w_{k}$ refers to the moving weight
and $G$/$K$ to the size of $F_{loc}$/$P_{loc}$.
To ease optimization, we further derive a proxy normalized loss quantity as:
\begin{equation}
    \mathbb{W}_{D} = \frac{\sum_{g=1}^{G}\sum_{k=1}^{K}f_{g,k}\delta_{g,k}}{\sum_{g=1}^{G}\sum_{k=1}^{K}f_{g,k}}.
\end{equation}
More details are given in the supplementary file.  
\subsubsection{Multi-modal attentive embedding}
After local alignment between point cloud and personalization input,
we further fuse the information across modalities. \saura{Motivated by the generality of cross-attention \cite{chen2021crossvit}, we leverage transformers \cite{vaswani2017attention} to fuse multi-modal features via cross-attention}.
%
Concretely, each Transformer module consists of a self-attention layer and a feed forward network.
We obtain the multi-modal panel embedding $F_{cm}$ via:
\begin{equation}
    F_{cm} = \mathcal{T}_{c}(P_{loc}, F_{loc},F_{loc}) \in \mathbb{R}^{K \times D}, 
\end{equation}
where 
we set the query as $P_{loc}$ and key/value both as $F_{loc}$ respectively.
As a result, each element in $F_{cm}$ is linked particularly with a specific panel class \saura{$p_{i}$ ($i \in \{1,..,K\}$).}
This \saura{disentanglement of panel specific features} facilitates the realization of panel personalization,
as each panel can be manipulated individually.

\begin{table*}[ht]

\centering
\caption{\textbf{Evaluation of personalized pattern design} on Panel IOU for 6 garment transfer cases. x$\shortrightarrow$y indicates before (x) and after (y) personalization. Abbreviations J: Jacket, JP: Jumpsuit, T: Tee, D: Dress, JS: Jacket Sleeveless. }
\setlength\tabcolsep{5pt}
\begin{tabular}{cc|ccccccccc}
\hline
\multicolumn{1}{l|}{\multirow{2}{*}{\bf{Modality}}} & \multirow{2}{*}{\bf{Method}} & \multicolumn{7}{c}{\bf{Panel IOU for personalized edits}}                                                                                                                                                                                                                                                                                                                                                                                                                      \\ \cline{3-9} 
\multicolumn{1}{l|}{}                          &                         & \multicolumn{1}{c|}{\bf{Case 1}}                   & \multicolumn{1}{c|}{\bf{Case 2}}                   & \multicolumn{1}{c|}{\bf{Case 3}}                   & \multicolumn{1}{c|}{\bf{Case 4}}                               & \multicolumn{1}{c|}{\bf{Case 5}}                   & \multicolumn{1}{c|}{\bf{Case 6}}                   &             \\ \hline
\multicolumn{2}{c|}{\bf{Combinations}}                                        & \multicolumn{1}{c|}{\bf{J to T}}                  & \multicolumn{1}{c|}{\bf{T to J}}                  & \multicolumn{1}{c|}{\bf{JP to D}}                  & \multicolumn{1}{c|}{\bf{D to JP}}                   & \multicolumn{1}{c|}{\bf{J to JS}}                 & \multicolumn{1}{c|}{\bf{JS to J}}      & \multicolumn{1}{c}{\bf{Avg}}                    \\ \hline
\multicolumn{1}{c|}{\multirow{2}{*}{Text}}     & Baseline                &\multicolumn{1}{c|}{0.32$\shortrightarrow$0.39}                        &\multicolumn{1}{c|}{0.27$\shortrightarrow$0.40}                        &\multicolumn{1}{c|}{0.18$\shortrightarrow$0.32}                                        &\multicolumn{1}{c|}{0.16$\shortrightarrow$0.31}                        &\multicolumn{1}{c|}{0.11$\shortrightarrow$0.32}                        &\multicolumn{1}{c|}{0.16$\shortrightarrow$0.43}    & \multicolumn{1}{c}{0.20$\shortrightarrow$0.36}                 \\ \cline{2-9} 
                \multicolumn{1}{c|}{}          & Ours                    &\multicolumn{1}{c|}{\cellcolor[HTML]{F6DDCC}0.46$\shortrightarrow$0.52} & \multicolumn{1}{c|}{\cellcolor[HTML]{F6DDCC} 0.41$\shortrightarrow$0.53} &\multicolumn{1}{c|}{ \cellcolor[HTML]{F6DDCC} 0.29$\shortrightarrow$0.51} &\multicolumn{1}{c|}{\cellcolor[HTML]{F6DDCC} 0.19$\shortrightarrow$0.48}&\multicolumn{1}{c|}{\cellcolor[HTML]{F6DDCC} 0.25$\shortrightarrow$0.54} &\multicolumn{1}{c|}{\cellcolor[HTML]{F6DDCC} 0.25$\shortrightarrow$0.60}  &\multicolumn{1}{c}  {\cellcolor[HTML]{F6DDCC} 0.31$\shortrightarrow$0.53}     \\ \hline
\multicolumn{1}{c|}{\multirow{2}{*}{Sketch}}   & Baseline                & \multicolumn{1}{c|}{0.29$\shortrightarrow$0.33}                        & \multicolumn{1}{c|}{0.24$\shortrightarrow$0.32}                         & \multicolumn{1}{c|}{0.15$\shortrightarrow$0.39}                                      & \multicolumn{1}{c|}{0.12$\shortrightarrow$0.37}                        & \multicolumn{1}{c|}{0.11$\shortrightarrow$0.34}                        & \multicolumn{1}{c|}{0.18$\shortrightarrow$0.40}   &    \multicolumn{1}{c}  {0.19$\shortrightarrow$0.36}                        \\ \cline{2-9} 
\multicolumn{1}{c|}{}                          & Ours                    & \multicolumn{1}{c|}{\cellcolor[HTML]{F6DDCC} 0.45$\shortrightarrow$0.52}                        & \multicolumn{1}{c|}{\cellcolor[HTML]{F6DDCC} 0.41$\shortrightarrow$0.52}                          & \multicolumn{1}{c|}{\cellcolor[HTML]{F6DDCC} 0.28$\shortrightarrow$0.51}                        & \multicolumn{1}{c|}{\cellcolor[HTML]{F6DDCC} 0.18$\shortrightarrow$0.46}                                      & \multicolumn{1}{c|}{\cellcolor[HTML]{F6DDCC} 0.25$\shortrightarrow$0.55}                        & \multicolumn{1}{c|}{\cellcolor[HTML]{F6DDCC} 0.27$\shortrightarrow$0.56}      &    \multicolumn{1}{c}  {\cellcolor[HTML]{F6DDCC} 0.31$\shortrightarrow$0.52}                                       \\ \hline
\end{tabular}

\label{tab:personalization}
\end{table*}

\begin{figure*}[t]
    \centering
    \includegraphics[scale=0.52]{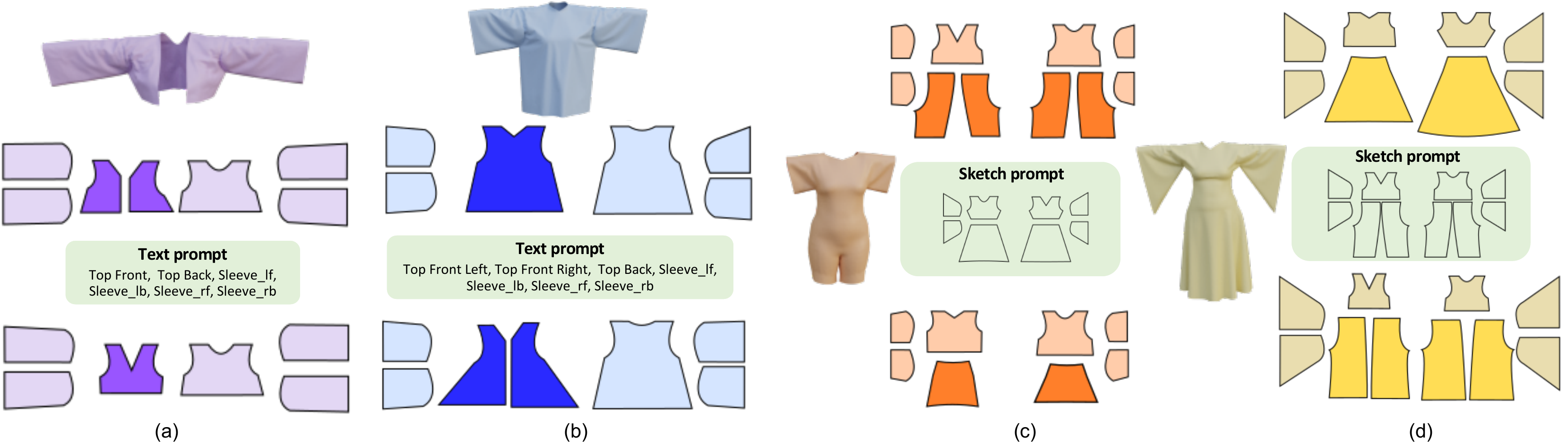}
    \caption{\ann{Examples of garment class transfer cases. Given a 3D source garment, we use the desired panel attributes to transfer it to the target garment class. (a) Case 1: Jacket to Tee by text prompt (target garment's panel classes), 
    (b) Case 2: Tee to Jacket by text prompt, 
    (c) Case 3: Jumpsuit to Dress by sketch prompt (target garment's average panel silhouettes), and (d) Case 4: Dress to Jumpsuit by sketch prompt. 
    The topology changes of panel are highlighted.
}}
    \label{fig:editing}
\end{figure*}
\subsection{Decoder: Panel mask prediction}
\saura{The panel-level decomposition of a given 3D garment forms the basis of our panel embedding based editing. This is similar in spirit with
shape generation by parts partitioning \cite{hertz2022spaghetti}.}
Given per-panel multi-modal embedding $F_{cm}$,
we predict the panel masks along with the stitching using a Transformer decoder $\mathcal{C}$ \cite{vaswani2017attention}.
Specifically, to exploit the panel's spatial information, we append positional encoding to $F_{cm} \in \mathbb{R}^{K \times D}$ with $K$ the number of panels. 
We set this embedding as the queries $Q$ of $\mathcal{C}$. 
We then apply self-attention on $F_{cm}$ for local interaction,
followed by cross-attention with the global feature $F_{global}$
to obtain the final panel-specific representation:
\begin{equation}
    F_{comp} = \mathcal{C}(F_{global};Q) \in \mathbb{R}^{K \times D}.
\end{equation}

\noindent {\bf Prediction heads} For efficiency, three lightweight heads are built to decode $F_{comp}$. 
The \emph{garment placement head} outputs the stitching information per panel. This is realized by training an MLP to output the rotation $\hat{r} \in \mathbb{R}^{M \times 3}$ and translation $\hat{t} \in \mathbb{R}^{M \times 3}$:
\begin{equation}\footnotesize
    \hat{r} = \sigma(Pool(W_{r}*F_{comp})), \hat{t} = \sigma(Pool(W_{t}*F_{comp}))
\end{equation}
where $W_{t}/W_{r} \in \mathbb{R}^{D \times 3}$ denotes the weights
$pool$ the average pooling operation. 
The \emph{panel confidence head} predicts the confidence of each panel. It is realized by another MLP followed by sigmoid operation:
\begin{equation}
    \hat{p}_{c} = \sigma(W_{p_{c}}*F_{comp}) \in \mathbb{R}^{D \times 1}
\end{equation}
where $W_{p_{c}}$ is the weight. 
The \emph{panel mask head} first unsamples each query in $F_{comp}$ to a fixed dimension binary mask and then applies sigmoid as 
\begin{equation}
   \hat{m}= \sigma(\psi(F_{comp})) \in \mathbb{R}^{H \times W}
\end{equation}
where $\psi(.)$ is a series of up-convolution followed by ReLU except the last layer.
\saura{More specifically, given a 3D garment point-cloud, we predict the 2D binary panel masks $\hat{m}$ per query
position using the mask head. We then obtain most confident ones (predicted by panel
confidence head $\hat{p}_{c}$, thresholded at 0.5).
We select top-$k$ panels among the most confident panel masks ($k=14$ is set
empirically). The placement head $\hat{r} / \hat{t}$ is used for draping the garment.}

\noindent \textbf{Panel mask smoothing} 
The sewing panels of garments typically present smooth outlines \cite{korosteleva2022neuraltailor}.
To exploit this prior, we reformulate the predicted panel mask as a closed piece-wise curve with every piece (edge) constrained to be Bezier spline. 
Given the predicted mask $\hat{m}_{i}$,
we estimate the 2D starting vertex $\hat{v}_{i}$ and the curvature $\hat{c}_{i}$ of a panel edge as:
\begin{equation}
    \{\hat{v}_{i},\hat{c}_{i}\} := \tau(\mathcal{B}(\hat{m}_{i}))
\end{equation}
where $\mathcal{B}(\cdot)$ denotes VGG net with an MLP classifier followed by tanh activation $\tau(\cdot)$.

\subsection{Model training}

%
We use the dataset $D_{train}$ to train the model. Model input includes the point set, and the user instruction in form of the text of the ground-truth panel classes (\ie, textual prompt) 
or the silhouettes of the ground-truth panels (\ie, visual prompt).
\saura{Since the user instruction is based on panel positions, we activate the ground-truth panel positions and set null (zero) vectors at other positions. At inference, this enables the removing-panel function.}
Model output includes the 2D panels and sewing patterns.
We adopt the MSE loss ($L_{place}$) for 
training the \emph{garment placement head}. 
To train the \emph{panel confidence head}, we assign each ground-truth panel position $j$ as $p_{h}(j) = 1$, otherwise 0. As panel position prediction is a multi-class multi-label problem, we use binary cross entropy loss ($\mathcal{L}_{conf}$). 
We train the \emph{panel mask head} using a binary cross-entropy loss ($\mathcal{L}_{mask}$). 
We assign each ground-truth panel position $j$ with its corresponding mask $m_{j} \in \mathbb{R}^{H \times W}$ to $p_{g}(j)$, otherwise an empty mask.
We use $L_1$ regression loss ($\mathcal{L}_{con}$) to estimate the curvature and vertex positions. 
We additionally use the local-association loss
($\mathcal{L}_{asso} = \mathbb{W}_{D}$)
to learn the cross-modal association. 
The overall objective is the sum of all above loss terms.

\subsection{Model inference and personalization}

Our model can support both standard and personalized pattern design.
The difference between the two settings 
lies in how to set the mask instruction $M$.
In the standard setting, 
given an unseen point cloud, we activate all the panels of $M$.
\saura{Although the training and testing garments share the same panel class set $p$, the panel combination that forms unseen garments is unknown.} Thus, for the textual prompt we use all $M$ panel classes ; For the visual prompt, we use the mean sketchRNN \cite{ha2017neural} embedding (\ie, the prototype of each of $M$ panels in all training classes). We output the top-$k$ most confident panel masks above a fixed threshold.
For {\em personalization}, we activate only the panels specified in the user's instruction mask \saura{(\ie, passing null vectors at other positions)} and output their 2D panels as shown in Fig~\ref{fig:decoder}(b).

\subsection{Garment stitching}
With the panel masks predicted,
we can further infer the stitching information for edge sewing across the panels. To that end, we design a StitchGraph module leveraging a GNN $\mathcal{G}$ \cite{yang2021sketchgnn}. 
Given a set of panel vertices $\hat{v}_{i}$, panel curvature $\hat{c}_{i}$ and placement information ($\hat{r}_{i}$, $\hat{t}_{i}$) of the panel mask $\hat{m}_{i}$,
we predict the stitching signal $\hat{s}$:
\begin{equation}
    \hat{s} = \mathcal{G}(\hat{v}_{i},\hat{c}_{i}) 
\end{equation}
where the value of ``1'' indicates the two edges stitched and ``0'' otherwise. 
We train $\mathcal{G}$ by a binary cross entropy loss $\mathcal{L}_{stitch}$.
This stitching signal coupled with panel placement information $\hat{r}_{i},\hat{t}_{i}$ is used for draping the garment on to the human body. \saura{Note, we follow the same stitching evaluation setting (\eg, no optimization of the stitching parameters in 3D space for different body shapes and sizes) as NeuralTailor \cite{korosteleva2022neuraltailor} for facilitating comparison.}

\section{Experiments}

\begin{table*}[t]
\small
\centering
\caption{\textbf{Evaluation of panel-prediction quality} on seen and unseen garment classes. M-L2: Mask L2 ; P-L2 : Panel L2; R-L2: Rotation L2; T-L2: Translation L2 . $\dagger$ represents orderless-LSTM.} 
\setlength{\tabcolsep}{8pt}
\begin{tabular}{@{}c|ccccc|ccccc@{}}
\toprule
                                   & \multicolumn{5}{c|}{\textbf{Seen classes}}                                                                                                                                                                                                      & \multicolumn{5}{c}{\textbf{Unseen classes}}                                                                                                                                                                                                     \\ \cmidrule(l){2-11} 
\multirow{-2}{*}{\textbf{Methods}}& \textbf{P-L2}                     & \textbf{\# Panels}                    & \textbf{\# Edges}                     & \textbf{R-L2}                       & \textbf{T-L2}           & \textbf{P-L2}                     & \textbf{\# Panels}                    & \textbf{\# Edges}                     & \textbf{R-L2}                       & \textbf{T-L2}                     \\ \midrule
Baseline-I                 & 3.92                                    &  \bf 99.9\%                                    &\bf 100.0 \%                                    & 0.06                             & 0.117                                       & 6.61                                    & 94.6\%                                     &  95.4\%                                     & 0.09                                    & 0.21                                    \\
Baseline-II                                                   & 4.3                                   & 99.4\%                                   &   99.7\%                                        & 0.08                                   & 1.46                                                             & 8.1                                   & 89.3\%                                   & 90.3\%                                   & 121                                   & 1.25                                   \\
LSTM                                                       & 2.71                                  & 99.8\%                                & 99.9\%                                & \bf 0.004                                 & 0.32                                                                 & 14.7                                  & 6.5\%                                 & 53.2\%                                & 0.17                                  & 6.75                                  \\
LSTM$^{\dagger}$                                                    & 2.87                                  & 99.4\%                                & 99.9\%                                & \textbf{0.004}                                 & 0.33                                                                   & 12.94                                 & 2.7\%                                 & 59.0\%                                & 0.16                                  & 7.18                                  \\
Neural-Tailor                                                      & \textbf{1.5}                                   & 99.7\%                                & 99.7\%                                & 0.04                                  & 1.46                                                         & 5.2                                   & 83.6\%                                & 87.3\%                                & 0.07                                  & 3.22                                  \\
\midrule
\textbf{Ours w/ Text}                      & \cellcolor[HTML]{FFFFDB}2.80  & \cellcolor[HTML]{FFFFDB}\textbf{99.9\%}  & \cellcolor[HTML]{FFFFDB}{99.9\%}  & \cellcolor[HTML]{FFFFDB}0.04  & \cellcolor[HTML]{FFFFDB}\textbf{0.04}    & \cellcolor[HTML]{FFFFDB}\textbf{4.20}  & \cellcolor[HTML]{FFFFDB}\textbf{99.9\%}  & \cellcolor[HTML]{FFFFDB}\textbf{99.8\%}  & \cellcolor[HTML]{FFFFDB}\textbf{0.05}  & \cellcolor[HTML]{FFFFDB}\textbf{0.05}  \\
\textbf{Ours w/ Sketch}                      & \cellcolor[HTML]{FFFFDB}2.91  & \cellcolor[HTML]{FFFFDB}\textbf{99.9\%}  & \cellcolor[HTML]{FFFFDB}{99.9\%}  & \cellcolor[HTML]{FFFFDB}0.05  & \cellcolor[HTML]{FFFFDB}{0.06}    & \cellcolor[HTML]{FFFFDB}\textbf{4.33}  & \cellcolor[HTML]{FFFFDB}\textbf{99.9\%}  & \cellcolor[HTML]{FFFFDB}\textbf{99.9\%}  & \cellcolor[HTML]{FFFFDB}\textbf{0.06}  & \cellcolor[HTML]{FFFFDB}\textbf{0.07}  \\

 \bottomrule 
\end{tabular}

\label{tab:main_tab}
\end{table*}

\begin{figure*}[t]
    \centering
    \includegraphics [width=\linewidth]{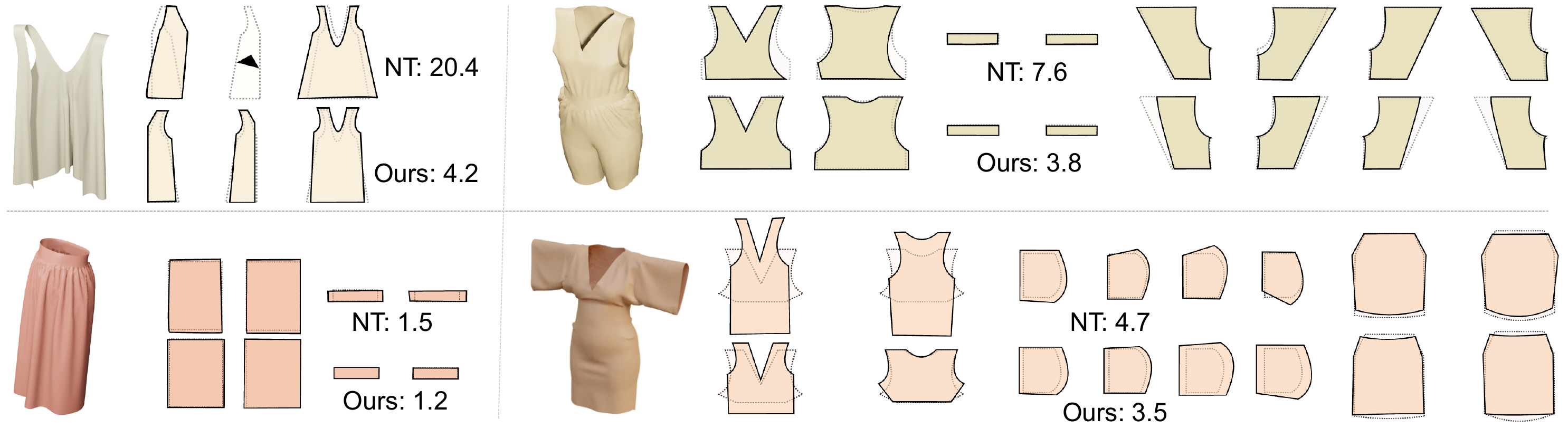}
    \caption{
    Comparing our method with NeuralTailor (\texttt{NT})
    on the unseen garment classes: ‘jacket sleeveless’, ‘skirt waistband’, ‘wb jumpsuit sleeveless’ and ‘dress’.
    {\em Metric}: the average Vertex L2 error.
    {\em Ground-truth}: dash thin lines.
    }
    \label{fig:main_viz}
\end{figure*}
\noindent \textbf{Dataset}
We evaluate the PersonalTailor on the 3D garments dataset with sewing patterns from \cite{korosteleva2021generating}. It contains 19 garment classes with $22,000$ 3D garment-sewing pattern pairs in total, covering the variations in t-shirts, jackets, pants, skirts, jumpsuits and dresses. 
There are 10627/722/729 samples for train/val/test
in the filtered version. 
Following NeuralTailor \cite{korosteleva2022neuraltailor}, the classes of panels are designed based on the panel's role and location around the body across all garment classes. For example, panels located around the back of human body are grouped in the ``back panels'' class. We follow the standard panel labels, data filtering and train/test splits of garment classes. There are 7 garment classes unseen to training and used for evaluation.

\noindent \textbf{Evaluation metrics}
We use the same evaluation metrics as in \cite{korosteleva2022neuraltailor}. 
We evaluate the accuracy in predicting the number of panels within
every pattern (\ie, \#Panels) and the number of edges within every panel
(\ie, \#Edges). To estimate the quality of panel shape predictions, we use the average distance (L2 norm) between the vertices of predicted and ground
truth panels with curvature coordinates converted to panel space,
acting as panel masks in this comparison (Panel L2). Similarly,
we report L2 normalized differences of predicted panel rotations
(Rot L2) and translations (Transl L2) with the ground truth. The
quality of predicted stitching information is measured by a mean
precision (Precision) and recall (Recall) of predicted stitches.

\noindent \textbf{Implementation details}
For language encoding, we use CLIP \cite{radford2021learning} pretrained encoder. 
For sketch encoding, we use SketchRNN \cite{yang2021sketchgnn}. 
We follow the training scheme as \cite{korosteleva2022neuraltailor}.
We set the maximum number of panels $M=23$.
There are $g=12/8$ garment classes in training/testing set. We set the feature dimension for text and the global embedding $D = 512$. 
Our model is trained for 250 epochs using Adam optimizer with learning rate of $10e-5$ and batch size of 15. 
The stitching GNN is trained for 50 epochs using SGD optimizer with learning rate of $10e-4$.
%
%
Specifically, 
it is trained by the predicted edges.
The inference threshold for panel mask head is set as 0.5 and top-$k$ is set as 14. The code will be made publicly available upon acceptance.


\subsection{Personalized pattern design evaluation}
\noindent \textbf{Setting}  To quantitatively evaluate the performance of personalization,
we conduct 6 garment class transfer cases (case 1\&2: Tee $\leftrightarrow$ Jacket, case 3\&4: Jumpsuit$\leftrightarrow$ Dress, case 5\&6: Jacket $\leftrightarrow$ Jacket Sleeveless)
under both text and sketch prompt. We define the \textit{Panel IOU}  metric as the mean of panel-wise IOUs between predicted panels of the source garment class and the average panels of the target garment class. Formally, we use the desired input prompts to transfer the source garment class to the target garment class. Then we compare the \textit{Panel IOU} before and after personalization against the target class panel attributes. 

\noindent \textbf{Baseline} Due to lacking of competing works or open-source alternatives, 
we created a personalization baseline by removing the prompt embedding and cross-modal embedding module (referred as \texttt{baseline}) from our PersonalTailor. 

\noindent \textbf{Quantitative results} 
The personalization results are reported in Tab.~\ref{tab:personalization}. It can be observed that (1) our method can achieve an average panel IOU of $53\%$ over 6 cases by text and $52\%$ by sketch, outperforming the baseline method by $13\%$/$16\%$ respectively. This is because the decoder of the baseline is randomly initialized lacking the semantic and structural information of the panel attributes. Thus, it has less personalization ability. (2) Our method yields a larger gain over the baseline before and after personalization under both text ($22\%$ \vs $16\%$) and sketch prompts  $21\%$ \vs $17\%$). 
This verifies our superior personalization ability.

\begin{figure}[t]

    \centering
    \includegraphics[scale=0.32]{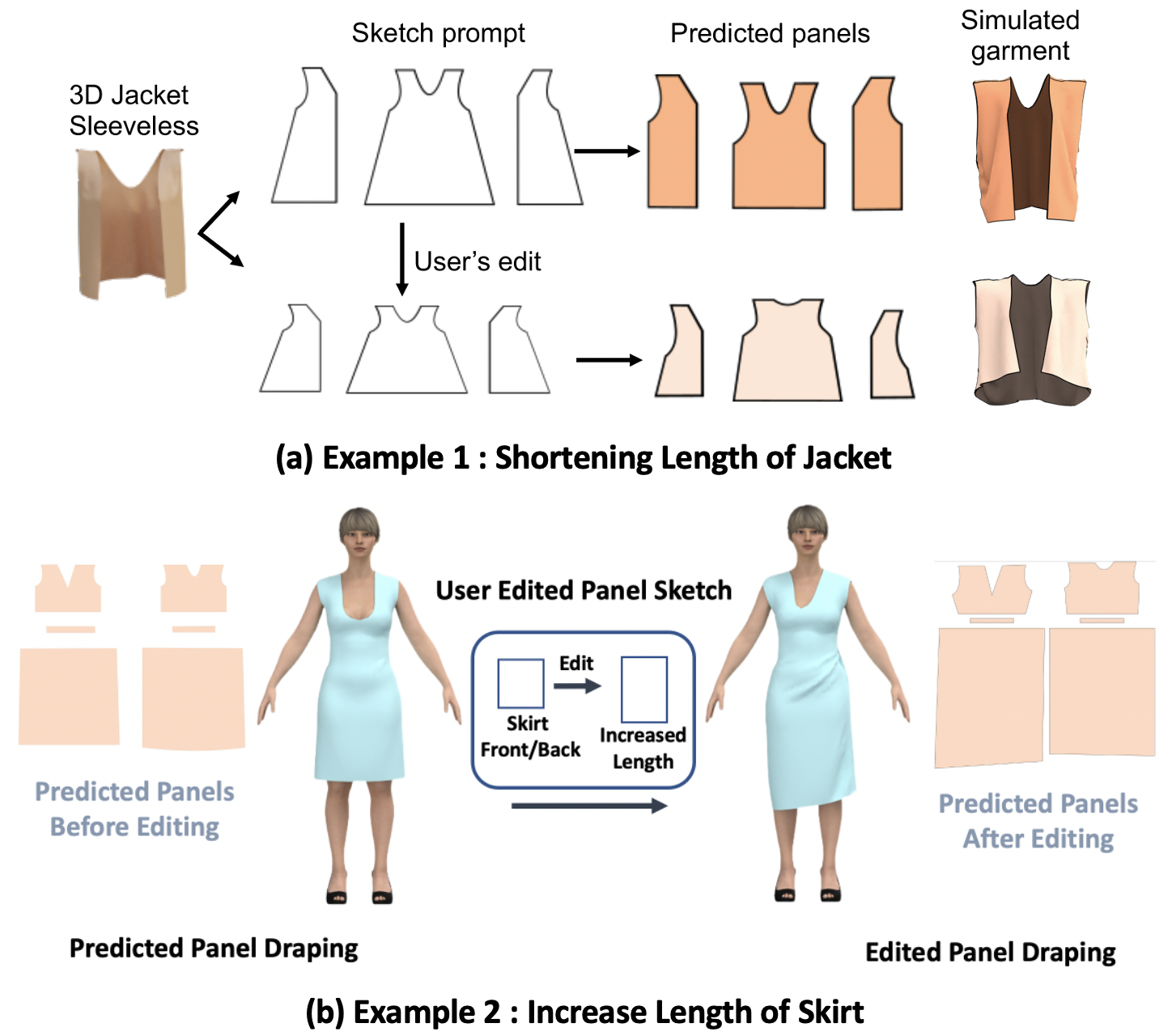}
    \caption{\ann{Illustration of fine-grained panel editing by sketch. Given a 3D garment and different users' sketches, our method can support fine-grained panel shape editing while preserving the intrinsic structure of the 3D garment.}}
  \label{fig:sketch_edit}
\end{figure}

\noindent \textbf{Qualitative/visual results}
We show the personalized garment transfer process of case 1\&2 by text prompt (target garment’s panel classes) in Fig.~\ref{fig:editing} (a,b), case 3\&4 by sketch prompt (target garment’s average panel silhouettes) in Fig.~\ref{fig:editing} (c,d).  Overall, it is shown that our method can support panel shape editing with complex topology changes from one garment class to another using personalized prompts, even for those unseen during training, \eg, Jumpsuit and Dress. Beyond topology change, it also supports adding 
new panels (Fig.~\ref{fig:teaser} (b)), removing panels (Fig.~\ref{fig:teaser} (c)), and creating a
new design 
(Fig.~\ref{fig:teaser} (e)).
We also observe that our method can achieve fine-grained panel shape editing by using sketch prompts. As shown in Fig.~\ref{fig:sketch_edit}, given a 3D jacket and different users' sketch prompts, our method can produce the panels aligned with the sketch's shape while preserving the intrinsic structure of the 3D shape. 


\subsection{Standard pattern design evaluation}

\noindent \textbf{Setting} In this setting, we evaluate the standard (non-personalized) pattern design.  
We follow the same setting and dataset splits as proposed in NeuralTailor \cite{korosteleva2022neuraltailor}. More specifically, we evaluate on two settings:
\ann{(1) Training with seen classes and evaluating on unseen data of those seen classes, \ie closed-set setting; 
(2) Training with seen classes and evaluating on unseen classes, \ie open-set setting.} 

\noindent \textbf{Competitors} We considered the following competitors for comparison: 
(a) a competitive garment pattern prediction method Neural Tailor on filtered data \cite{korosteleva2021generating}, 
(c) an LSTM \cite{graves2012long} based garment pattern prediction ,
(d) an orderless LSTM \cite{yazici2020orderless} based garment pattern prediction, 
(e) \textit{Baseline-I} we created using GCN encoder and CNN decoder, 
(f) \textit{Baseline-II} we created using PointTransformer encoder \cite{zhao2021point} and Transformer decoder \cite{vaswani2017attention} with random initialized queries. 

\noindent \textbf{Results} The results are reported in Tab.~\ref{tab:main_tab}. 
\textbf{(I) Closed-set settings:} 
The performance of some metrics (\eg, Edges/Panels) has almost saturated.
In particular, NeuralTailor has the best Panel L2 result, indicating that learning vertex is better in the closed set than mask prediction.
However, our PersonalTailor achieves the best translation prediction, suggesting the importance of global information.
%
\textbf{(II) Open-set settings:} 
Our method achieves the state-of-the-art in all the metrics, surpassing the competitors by a large margin. This indicates the superiority of \ann{personalized prompts} in open-set generalization.

\paragraph{Qualitative results} 
We present qualitative results on unseen garments. 
It can be observed from Fig.~\ref{fig:main_viz} that our PersonalTailor predicts more accurate panels over the prior art NeuralTailor \cite{korosteleva2022neuraltailor}, due to our multi-modal embedding-based design enabling the prompt bring in additional semantic information about the garment's shape.  
We also show in Fig. \ref{fig:ptailor_main} that PersonalTailor works well with both text and sketch prompts.

\begin{figure}[t]
    \centering
    \includegraphics[scale=0.35]{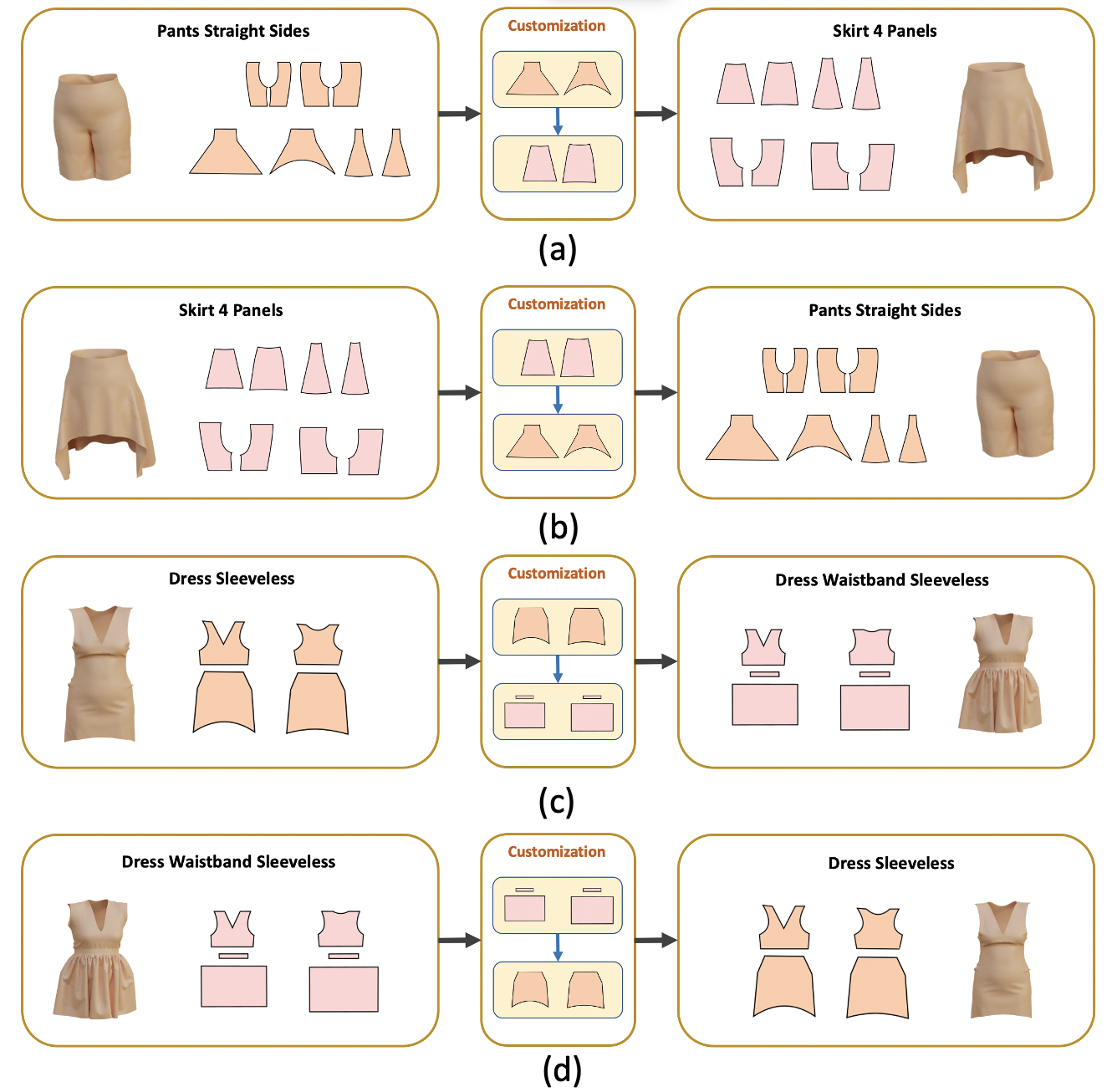}
    \caption{\textbf{Examples of garment personalization} 
    %
    from (a) Pant Straight sides to Skirt 4 Panels, (b) Skirt 4 Panels to Pant Straight Sides, (c) Dress Sleeveless to Dress Waistband Sleeveless, (d) Dress Waistband Sleeveless to Dress Sleeveless respectively. }
    \label{fig:customized}
\end{figure}

\subsection{Stitching prediction} 

We evaluate the stitching module design
by comparing with NeuralTailor \cite{korosteleva2022neuraltailor}.
As shown in Tab.~\ref{tab:stitch}, 
our GNN based design is clearly superior particularly
for unseen garment classes.
This validates the efficacy of our exploiting the structural information of panels.
We also show that using the edge vectors of ground-truth panels
for training is inferior than using the predicted for both methods,
as the former introduces some inconsistency with model inference.


\begin{table}[h]
\caption{\textbf{Evaluation of stitching prediction} on both seen and unseen garment classes.
$^*$: Trained by the edges of GT panels.
}
\label{tab:stitch}
\resizebox{\columnwidth}{!}{
\begin{tabular}{c|cc|cc}
\hline
\multirow{2}{*}{\textbf{Method}}                 & \multicolumn{2}{c|}{\textbf{Seen classes}}     & \multicolumn{2}{c}{\textbf{Unseen classes}}   \\ \cline{2-5}
                & \textbf{Precision}       & \textbf{Recall}          & \textbf{Precision}       & \textbf{Recall}          \\ \hline
NeuralTailor$^*$ \cite{korosteleva2022neuraltailor}  & 96.6\%          & 88.6\%          & 75.3\%          & 60.6\%          \\ \hline
NeuralTailor \cite{korosteleva2022neuraltailor} & 96.3\%          & 99.4\%          & 74.7\%          & 83.9\%          \\ \hline
Ours$^*$            & 74.9\%          & 65.0\%          & 76.8\%          & 73.0\%          \\ \hline
Ours          & \cellcolor[HTML]{F6DDCC}\textbf{99.9\%} & \cellcolor[HTML]{F6DDCC}\textbf{99.9\%} & \cellcolor[HTML]{F6DDCC}\textbf{85.8\%} & \cellcolor[HTML]{F6DDCC}\textbf{84.2\%} \\ \hline
\end{tabular}
}
\label{stitchingres}
\end{table}

\begin{figure*}
    \centering
    \includegraphics[scale=0.43]{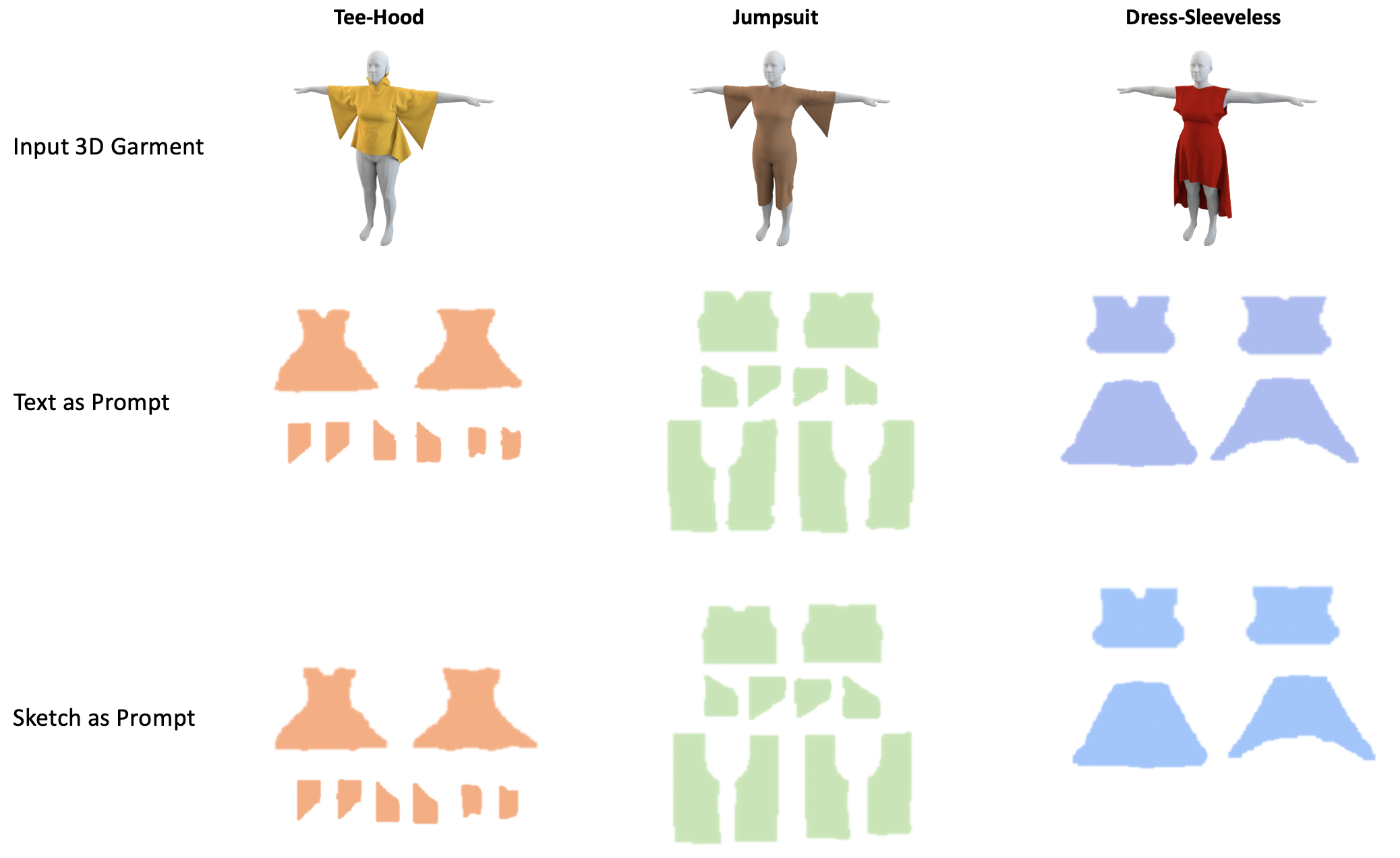}
    \caption{\textbf{Examples of PersonalTailor's output (unrefined)}
    It is shown that our method works similarly well with text and sketch/visual prompts.
    }
    \label{fig:ptailor_main}
\end{figure*}
\subsection{Ablation studies}
We conduct ablation studies to provide insights into each
component with text prompt. \saura{More in-depth ablations are provided in the \texttt{Supplementary}.}

\noindent \textbf{Impact of cross-modal alignment}
We evaluate the importance of cross-modal alignment.
To that end, we consider two down-stripped designs:
{\bf(1)} Removing the cross modal local association block (CMLA);
{\bf(2)} Removing the multi-modal transformer (MMT).
As shown in Tab.~\ref{tab:mmemb},
we find that without CMLA, a significant drop in the Panel L2 metric 
occurs, suggesting the importance of resolving the domain gap between the point cloud and semantic information from the prompt.
It is also shown that MMT is useful in terms of 
fusing information.

\begin{table}[h]
\centering
\small
\caption{Impact of multi-modal embedding.
CMLA: Cross modal local association;
MMT: Multi-modal transformer.
}
\label{tab:mmemb}
\setlength{\tabcolsep}{3pt}
\begin{tabular}{c|cc|cc}
\hline
\multirow{2}{*}{\textbf{Model}} & \multicolumn{2}{c|}{\textbf{Seen}}     & \multicolumn{2}{c}{\textbf{Unseen}}    \\ \cline{2-5} 
                                & \textbf{Panel L2} & \textbf{\# panels} & \textbf{Panel L2} & \textbf{\# panels} \\ \hline
 Ours                            & \cellcolor[HTML]{F6DDCC}\textbf{2.80}                & \cellcolor[HTML]{F6DDCC} \textbf{99.9\%}                 & \cellcolor[HTML]{F6DDCC}\textbf{4.20}               & \cellcolor[HTML]{F6DDCC}\textbf{99.9\%}  \\ \hline
w/o CMLA                         & 5.51               & 99.8\%                & 6.5                & 99.2\%                   \\
w/o MMT                         & 4.42                &  99.9\%                  & 5.32     
& 99.6\%                  \\ 
\hline

\hline

\end{tabular}
\end{table}


\noindent \textbf{Impact of point-cloud encoders} We evaluate our point-cloud encoder design including 
global encoder (GE) and local encoder (LE).
As shown in Tab.~\ref{tab:ptc}, we see that 
{\bf(1)} GE is useful particularly for unseen garment classes.
This is because global information plays an important role in estimating the cutting pattern and guiding the panel-mask decoder prediction.
{\bf (2)} LE brings in further performance gain
due to extra part-level information introduced.
%

\begin{table}[t]
\centering
\small
\setlength{\tabcolsep}{3pt}
\caption{Impact of global and local point cloud encoders.
LE: Local Encoder; GE: Global Encoder.}
\label{tab:ptc}
\begin{tabular}{cc|cc|cc}
\hline
\multicolumn{2}{c|}{\textbf{Model} }      & \multicolumn{2}{c|}{\textbf{Seen}} & \multicolumn{2}{c}{\textbf{Unseen}} \\
\hline
 \textbf{LE}     & \textbf{GE}                       & \textbf{Panel L2}    & \textbf{\# Panels}      & \textbf{Panel L2}     & \textbf{\# Panels}     \\ \hline
\xmark          & \cmark                         & 3.20          & 99.4\%            & 4.90               & 95.2\%            \\
\cmark         & \xmark         & 3.30          & 99.9\%            & 5.51               & 96.3\%      \\
\hline
 \cmark        & \cmark    & \cellcolor[HTML]{F6DDCC}\textbf{2.80} & \cellcolor[HTML]{F6DDCC}\textbf{99.9\%} & \cellcolor[HTML]{F6DDCC}\textbf{4.20}  & \cellcolor[HTML]{F6DDCC}\textbf{99.9\%} \\ \hline
\end{tabular}
\end{table}

\noindent \textbf{Design choice of panel decoder}
We evaluate more choices of panel mask decoder
including (1) a CNN and (2) a Transformer decoder without positional embedding (Trans. w/o PE). 
We observe in Tab.~\ref{tab:dec} that (1) CNN is least performing for instruction based mask prediction as it loses out on the panel prediction performance due to lack of interaction among the panels. 
(2) Positional encoding is important as it predicts position specific masks.

\begin{table}[h]
\centering
\setlength{\tabcolsep}{3pt}
\caption{Ablation on the design choice of decoder.
Trans.: Transformer; PE: Positional Encoding.
}
\label{tab:dec}
\begin{tabular}{c|cc|cc}
\hline
\multirow{2}{*}{\textbf{Model}}       & \multicolumn{2}{c|}{\textbf{Seen}} & \multicolumn{2}{c}{\textbf{Unseen}} \\ \cline{2-5} 
                             & \textbf{Panel L2}    & \textbf{\# Panels}      & \textbf{Panel L2}     & \textbf{\# Panels}     \\ \hline
CNN                          & 5.49          & 93.2\%          & 6.92           & 91.7\%          \\
Trans. w/o PE        & 3.30          & 99.9\%            & 5.61           & 96.1\%    \\
\hline
Ours & \cellcolor[HTML]{F6DDCC}\textbf{2.80} & \cellcolor[HTML]{F6DDCC}\textbf{99.9\%} & \cellcolor[HTML]{F6DDCC}\textbf{4.20}  & \cellcolor[HTML]{F6DDCC}\textbf{99.9\%} \\ \hline
\end{tabular}
\end{table}


\subsection{In-the-wild garment pattern design}
\begin{figure}[h]
    \centering
    \includegraphics[scale=0.28]{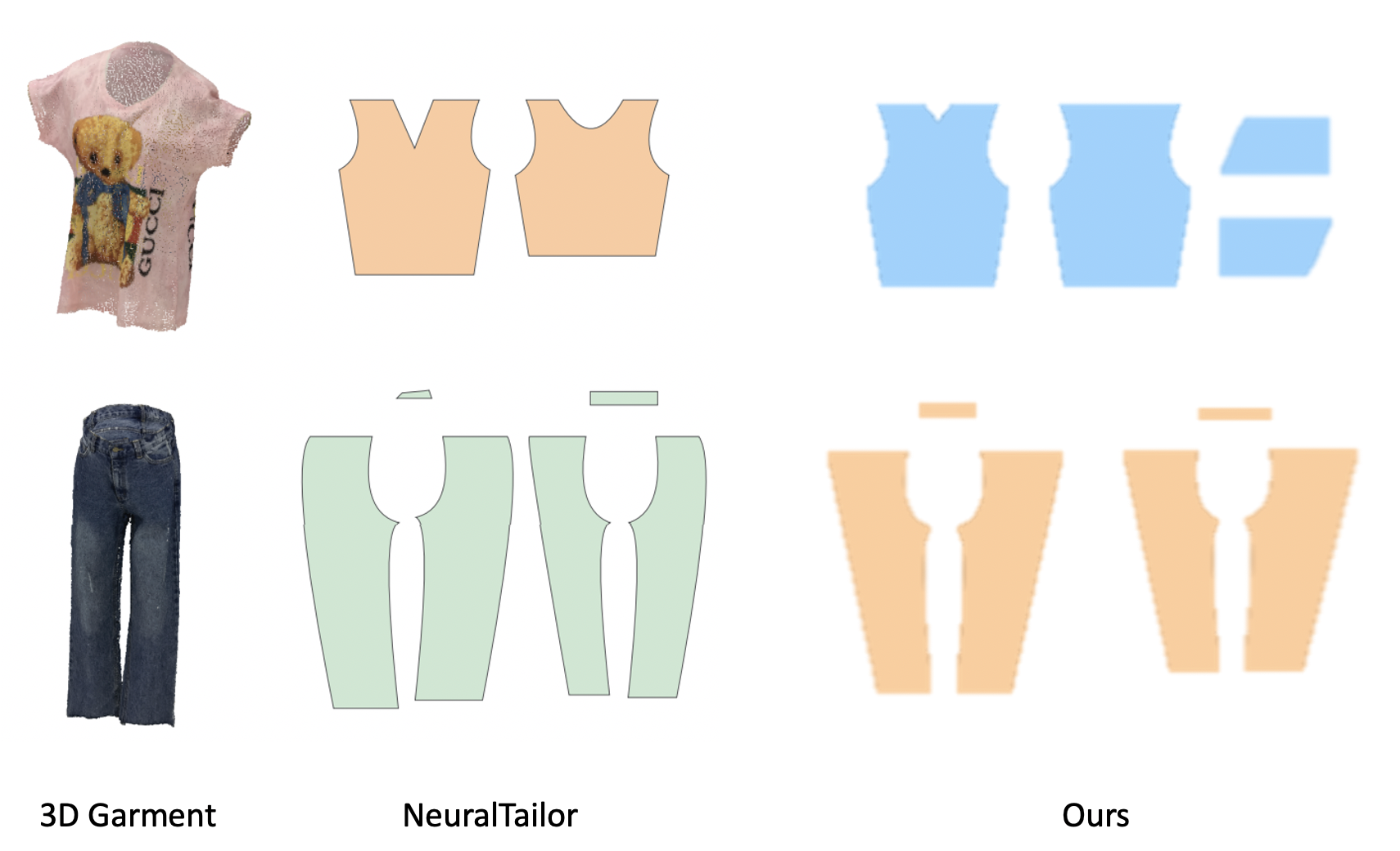}
    \caption{\textbf{In-the-wild garment evaluation} Sewing patterns predicted by NeuralTailor \cite{korosteleva2022neuraltailor} and our PersonalTailor on examples from DeepFashion3D \cite{zhu2020deep}.}
    \label{fig:wild}
\end{figure}

For more extensive evaluation, we qualitatively test on the garment captures from DeepFashion3D dataset \cite{zhu2020deep}.
We observe in Fig.~\ref{fig:wild} that our model makes better panel predictions than NeuralTailor \cite{korosteleva2022neuraltailor}. For example, NeuralTailor fails to perceive the sleeves
with the T-shirt (unseen to model training), whilst our model succeeds.
In the case of jeans, our model gives better panel uniformity than \cite{korosteleva2022neuraltailor}.

\section{Human Evaluation}

\begin{figure}[h]
    \centering
    \includegraphics[scale=0.38]{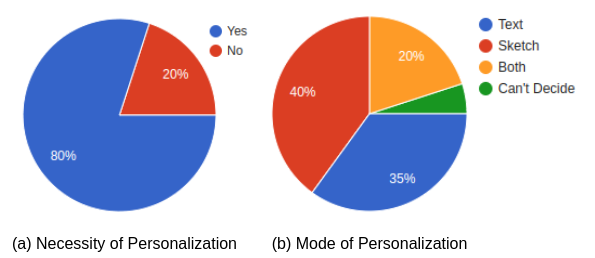}
    \caption{\textbf{Votes on garment personalization.}}
    \label{fig:hstud1}
\end{figure}

\begin{figure}
    \centering
    \includegraphics[scale=0.23]{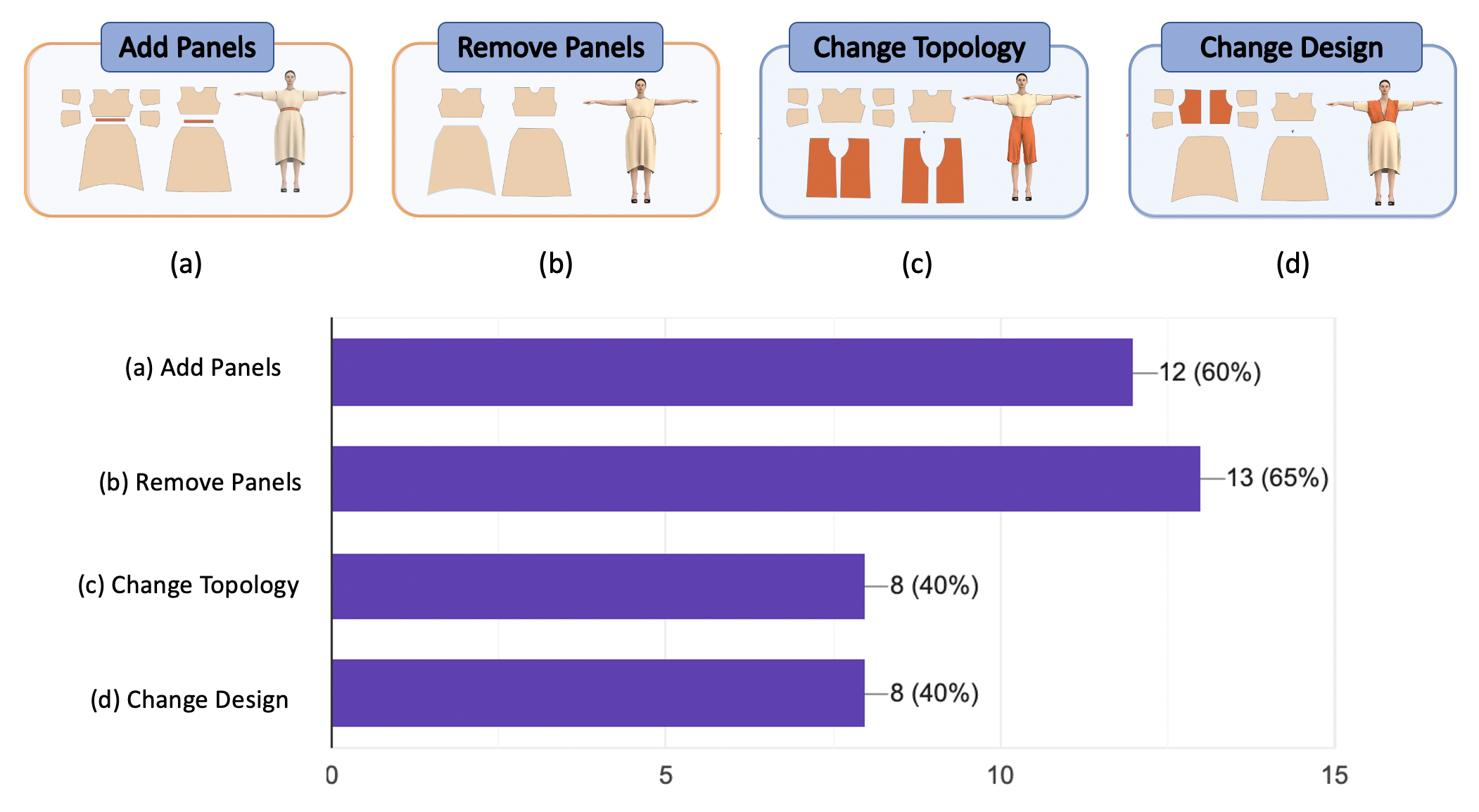}
    \caption{\textbf{Votes on garment personalization functions.}}
    \label{fig:hstud2}
\end{figure}

In additional to benchmark based assessment,
we further provide human evaluation with a thoughtful user study.
In particular, we approached 20 professional tailors to request their 
preference on the significance of personalizing garment pattern design.
In particular, we asked them two questions: 
(1) If garment personalization is necessary? 
(2) Which prompt (text or sketch) is preferred?
As shown in Fig.~\ref{fig:hstud1}, 80\% of tailors consider 
automated personalization to be useful, as it saves them time in production and the cost of business.
Besides, 40\% prefer sketch over text for instruction,
whilst 5\% are concerned with sketch to be only for professional designers.
We also collected the tailors feedback on the functions of (a) adding new garment panels, (b) removing garment panels, (c) changing dress topology, and (d) creating new designs. As shown in Fig.~\ref{fig:hstud2}, removing garment panels is most popular, which is supported by our proposed method. 


\section{Limitations and Future Work}
We presented a personalized 2D pattern design method for 3D garments,
featuring editing capabilities. Our model is 2D panel-aware, requires no panel annotation, and leverages the Transformer architecture to
form globally coherent 2D patterns of varied topology. The network was designed to allow editing at an interactive rate, where, as demonstrated, the user can interact with the model using a simple instruction (refer to Fig 2). However, one limitation is a relatively small set of panels and
garments, lacking multiple test domains for evaluation. Our mask based panel design cannot model complicated 2D patterns like pleats or darts which is another limitation.  As this is an under-studied area, many challenges (3D optimization, fitting to different body structures etc.) still exist for future work.

\section{Conclusion}
This paper presented PersonalTailor, the first personalized 2D pattern design method supporting the user's instruction (\eg, language or sketch) for design pattern from the 3D point cloud garment. It is characterized by multi-modal panel embedding and panel-mask prediction using an encoder-decoder framework. 
Extensive experiments show that our method excels over existing alternatives on both standard and personalized garment pattern design settings.
Except fine-grained component-level ablation analysis,
a dedicated user study was also conducted, indicating the significance of our research in terms of both problem formulation and model design.


\paragraph{Acknowledgements} This research was conducted at the absence of any commercial/financial/personal relationships that could be construed as a potential conflict of interest. The authors would like to thank Li Jiaying for helping with conducting the human evaluation. The authors also would like to thank Wamiq Reyaz Para for providing insightful suggestions with the architecture design. 
{
    \small
    \bibliographystyle{ieeenat_fullname}
    \bibliography{main}
}
\clearpage
\setcounter{page}{1}
\maketitlesupplementary

\section{Rationale}
\label{sec:rationale}
Having the supplementary compiled together with the main paper means that:
\begin{itemize}
\item The supplementary can back-reference sections of the main paper, for example, we can refer to \cref{sec:intro};
\item The main paper can forward reference sub-sections within the supplementary explicitly (e.g. referring to a particular experiment); 
\item When submitted to arXiv, the supplementary will already included at the end of the paper.
\end{itemize}
To split the supplementary pages from the main paper, you can use \href{https://support.apple.com/en-ca/guide/preview/prvw11793/mac#:~:text=Delete%20a%20page%20from%20a,or%20choose%20Edit%20%3E%20Delete).}{Preview (on macOS)}, \href{https://www.adobe.com/acrobat/how-to/delete-pages-from-pdf.html#:~:text=Choose%20%E2%80%9CTools%E2%80%9D%20%3E%20%E2%80%9COrganize,or%20pages%20from%20the%20file.}{Adobe Acrobat} (on all OSs), as well as \href{https://superuser.com/questions/517986/is-it-possible-to-delete-some-pages-of-a-pdf-document}{command line tools}.

\end{document}